\pdfoutput=1

\documentclass[11pt]{article}

\usepackage[final]{acl}

\usepackage{times}
\usepackage{latexsym}

\usepackage[T1]{fontenc} 

\usepackage[utf8]{inputenc}

\usepackage{microtype}

\usepackage{inconsolata}

\usepackage{graphicx}
\usepackage{lipsum}
\usepackage{enumitem}
\usepackage{multirow}
\usepackage{fancyvrb}
\usepackage{listings}
\usepackage{booktabs}  
\usepackage{makecell}  
\usepackage{array}  
\usepackage{xcolor}  
\usepackage{longtable}  
\usepackage{graphicx}  
\usepackage{cleveref}

\crefformat{section}{\S#2#1#3}
\crefformat{subsection}{\S#2#1#3}
\crefformat{subsubsection}{\S#2#1#3}

\title{Can LLMs Help Uncover Insights about LLMs? \\ A Large-Scale, Evolving Literature Analysis of Frontier LLMs}

\author{Jungsoo Park$^{1}$\quad Junmo Kang$^{1}$\quad Gabriel Stanovsky$^{2}$\quad Alan Ritter$^{1}$ \\
  Georgia Institute of Technology$^{1}$ \quad The Hebrew University of Jerusalem$^{2}$\\
  \texttt{jpark3272@gatech.edu} \quad \texttt{junmo.kang@gatech.edu} \\
  \texttt{gabriel.stanovsky@mail.huji.ac.il} \quad
  \texttt{alan.ritter@cc.gatech.edu} \\}

\newcommand
{
\datasetname
}{
\textsc
{LlmEvalDB}}

\begin{document}
\maketitle
\begin{abstract}
The surge of LLM studies makes synthesizing their findings challenging. 
Analysis of experimental results from literature can uncover important trends across studies, but the time-consuming nature of manual data extraction limits its use.
Our study presents a semi-automated approach for literature analysis that accelerates data extraction using LLMs.
It automatically identifies relevant arXiv papers, extracts experimental results and related attributes, and organizes them into a structured dataset, \datasetname.
We then conduct an automated literature analysis of frontier LLMs, reducing the effort of paper surveying and data extraction by more than 93\% compared to manual approaches.
We validate \datasetname~by showing that it reproduces key findings from a recent manual analysis of Chain-of-Thought (CoT) reasoning and also uncovers new insights that go beyond it, showing, for example, that in-context examples benefit coding \& multimodal tasks but offer limited gains in math reasoning tasks compared to zero-shot CoT.
Our automatically updatable dataset enables continuous tracking of target models by extracting evaluation studies as new data becomes available. 
Through \datasetname~and empirical analysis, we provide insights into LLMs while facilitating ongoing literature analyses of their behavior.
\end{abstract}

\section{Introduction}

\begin{figure}[t!]
    \centering
    \includegraphics[width=0.95\columnwidth]
    {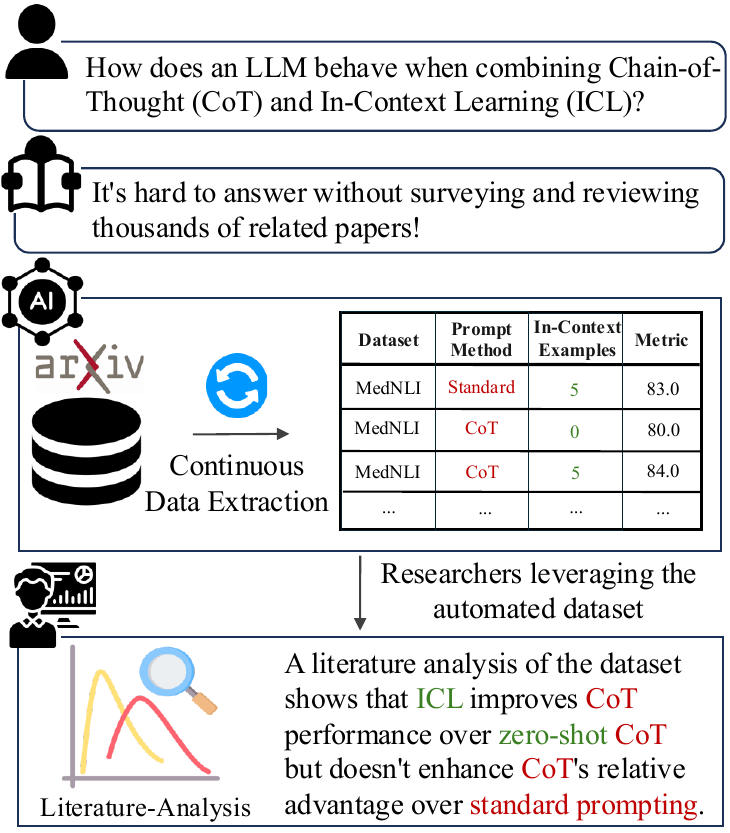}
    \caption{The diagram of our semi-automated literature analysis process and a key finding derived from data automatically extracted from the arXiv database.}
    \label{fig:teaser}
\end{figure}

The rapid advancement of Large Language Models (LLMs)~\citep{brown2020language, chowdhery2022palm, achiam2023gpt, touvron2023llama, team2023gemini, anthropic@claude} has led to a proliferation of empirical studies. 
Recent \textit{``surveys of surveys''}~\citep{ABigSurvey, ABigSurveyOfLLMs} highlight the overwhelming growth of LLM publications, surpassing what individual researchers can manually review.
The growing number of evaluations—each using different models, datasets, and prompting configurations—makes it challenging to analyze and synthesize findings across studies, underscoring the urgent need for automated analysis.

A manual literature analysis
that examines data from multiple independent studies to identify overall trends, offers a solution~\citep{reiter2018structured, hupkes2023taxonomy}.  
However, conducting such an analysis is time-consuming~\citep{yun2023appraising, yun2024automatically}.
For instance, as shown in Fig.~\ref{fig:teaser}, understanding the effects of combining Chain-of-Thought (CoT) with in-context examples requires identifying relevant papers, extracting specific experimental results, and aggregating the data.
Furthermore, analyses based on existing studies are likely to become quickly outdated in a fast-moving field, motivating the need for automated analyses that can be continuously updated as new results are shared on preprint servers such as arXiv.org.

In this study, we conduct the most extensive quantitative literature analysis of frontier LLMs to date. 
Our dataset, \datasetname, comprises 18,127 experimental records extracted from 1,737 papers, along with 359 referenced papers describing datasets used in these experiments. 
\datasetname~can also be dynamically updated with minimal human effort to incorporate results from new models or studies as they are published.
This comprehensive dataset enables us to present new insights into LLM performance, where previous studies on extracting leaderboards from machine learning papers have primarily focused on improving data extraction accuracy \citep{kardas2020axcell, yun2024automatically}.

To achieve this, we experiment with an automated approach to data extraction that efficiently processes research literature. 
This approach uses an LLM to scan the arXiv database, identify relevant papers with experimental results on target models, and extract experimental results along with pertinent attributes. 
This reduces the manual effort needed for surveying and extraction by more than 93\% compared to expert manual data extraction.
The approach employs schema-driven extraction~\citep{bai2023schema} and context augmentation from paper contents to capture essential attributes for a comprehensive understanding of evaluation results. 
Additionally, the process creates dataset descriptions that summarize key characteristics of the dataset associated with the evaluation record, enhancing \datasetname's potential utility in literature analysis. 

Our analysis shows that \datasetname~can efficiently replicate a previous manual analysis~\citep{sprague2024cot}, which found that CoT reasoning primarily improves performance on mathematical and symbolic reasoning benchmarks. 
We also identify three novel insights from the dataset on prompting configurations. 
First, in-context learning (ICL) enhances coding \& multimodal tasks but offers limited benefits for mathematical tasks compared to a zero-shot setup. 
Second, our qualitative analysis reveals specific characteristics of datasets that tend to reduce the effectiveness of CoT prompting and ICL.
Third, we find that CoT prompting with demonstrations—commonly implemented through ICL—typically yields better performance than CoT without demonstrations.
However, the relative improvement of CoT over standard direct prompting remains stable regardless of the number of demonstrations used (i.e., it holds true in both zero-shot and few-shot settings).
We release our dataset and source code to support the ongoing automated integration of new findings in the literature.
\footnote{The code is available at \url{https://github.com/JJumSSu/meta-analysis-frontier-LLMs}, and the dataset can be accessed at \url{https://huggingface.co/datasets/jungsoopark/LLMEvalDB}.}


 
\begin{figure*}[t!]
\centering
\includegraphics[width=0.9\textwidth]{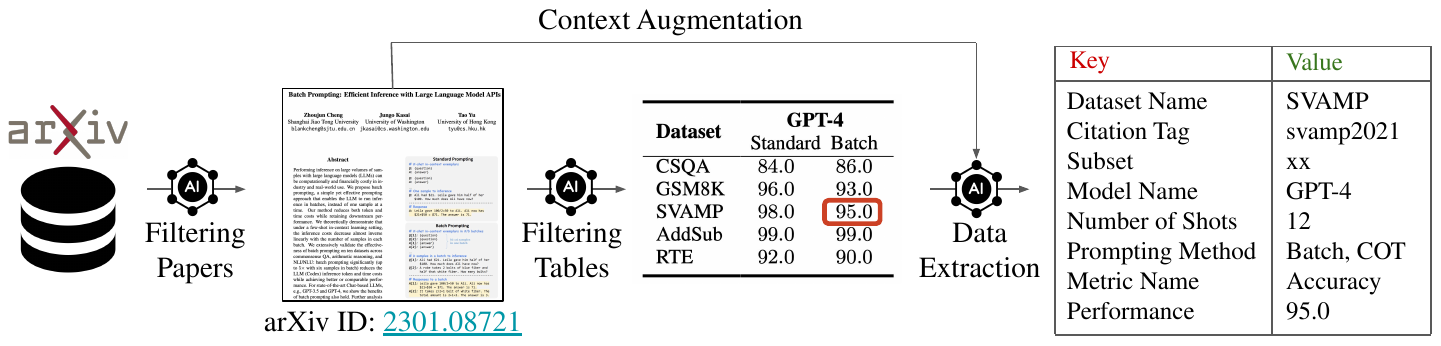}
\caption{Data extraction pipeline overview with an extracted example from arXiv paper~\citep{cheng2023batch}. Target attributes are identified and extracted from the table, augmented with paper content.}
\label{fig:data_extraction}
\end{figure*}

\begin{figure*}[t!]
\centering
\includegraphics[width=0.9\textwidth]{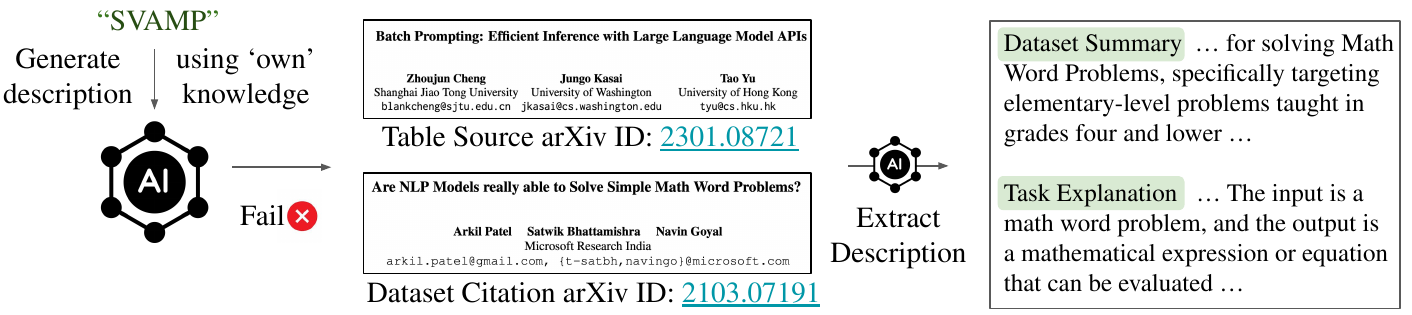}
\caption{Dataset description generation pipeline overview using the example of "SVAMP" dataset~\citep{patel2021nlp}. Since LLM lacked knowledge of the given dataset, the model references the original dataset's arXiv source retrieved through extracted BibTex. If confident, however, the LLM's generated descriptions will be used.}
\label{fig:data_description}
\end{figure*}

\section{Data Extraction}
\label{section:approach}

Automated data extraction is essential for scaling and improving literature analysis efficiency. 
This section outlines our pipeline for extracting experimental results and model attributes from arXiv sources, which are used to construct \datasetname.
We define target models and attributes (\cref{subsection:attributes}) and introduce our LLM-powered extraction process, which includes three stages: preprocessing \& filtering (\cref{subsection:preprocessing}), extraction \& augmentation (\cref{subsection:extraction}), and description generation (\cref{subsection:generation}).
Unless stated otherwise, we utilize GPT-4o \citep{hurst2024gpt} throughout the pipeline.

\subsection{Defining Target Models and Attributes}
\label{subsection:attributes}

\paragraph{Target Models} 

We analyze four leading proprietary models in the NLP/AI field: GPT-4~\citep{achiam2023gpt}, GPT4-o~\citep{hurst2024gpt}, Claude3 Opus~\citep{anthropic@claude}, and Gemini 1.0 Pro~\citep{team2023gemini}. 
Our analysis focuses on proprietary LLMs accessible via APIs that are not possible to fine-tune.\footnote{GPT-4o and Gemini 1.0 Pro became available for fine-tuning during our study period, but we used rule-based heuristics to filter out limited fine-tuned results.} We chose to exclude fine-tuned models from our study because comparing diverse fine-tuning methods would make controlled cross-study comparisons much more challenging. 
We also exclude recently released advanced reasoning models (GPT4-o1~\citep{GPT4o1} and Deepseek-R1~\citep{guo2025deepseek}) due to very limited published results to date. 
In the future, our pipeline can readily extract experimental data for these models along with those of other target models, as more studies are published. 
For further discussion, see \cref{sec:limitation}.

\paragraph{Target Attributes} To enable thorough analysis across different studies, we extract various performance-related fields from proprietary models.
The targeted attributes include: \textit{Dataset Name}, \textit{Subset}, \textit{Model Name}, \textit{Prompting Method}, \textit{Number of Demonstrations}, \textit{Metric Name}, and \textit{Performance}. 
Note that the subset refers to any division of the dataset, such as a subtask, domain, split, or language pair.
For instance, as shown in Fig.\ref{fig:data_extraction}, the arXiv paper with ID \textit{2301.08721}~\citep{cheng2023batch} presents experimental results for \textit{GPT-4}~\citep{achiam2023gpt} on the \textit{SVAMP} dataset~\citep{patel2021nlp}. 
The evaluation used \textit{Batch Prompting with CoT}~\citep{wei2022chain, cheng2023batch} and \textit{12 in-context samples}~\citep{brown2020language}, achieving an \textit{accuracy} score of \textit{95.0}.

\subsection{Preprocessing \& Filtering}
\label{subsection:preprocessing}

To extract the experimental results of target models and attributes, we used arXiv sources from January 2023 to December 2024, focusing on machine learning papers (cs.CV, cs.AI, cs.CL, cs.LG). 
arXiv LaTeX sources preserve structure better than parsed PDFs, enabling easier dataset extraction—especially from tables—using simple regex, while PDFs require complex tools. 
Hence, following prior work~\citep{kardas2020axcell, bai2023schema, wang2024charxiv}, we adopt arXiv LaTeX sources for data extraction, as over 90\% of arXiv papers are submitted in LaTeX format~\citep{frankston2024html}.

We downloaded LaTeX source files and applied regex-based methods to extract structured data like tables.
Before extraction, we filtered tables to reduce time and API costs by selecting those with target model results.
We also concentrated on leaderboard tables presenting benchmark results following~\citet{kardas2020axcell} and used Llama3.1-70B-Instruct \citep{dubey2024llama} for the filtering.

\subsection{Extraction \& Augmentation}
\label{subsection:extraction}

Based on the selected tables, we extract records from tables and paper contents. 
Using schema-driven extraction, we obtain records matching target attributes from tables, focusing only on relevant model data to save costs while maintaining accuracy. 
This selective approach significantly reduces API costs compared to extracting full results from each table~\citep{bai2023schema}. 
After table extraction, we augment records with context from the entire paper, adding experimental specifics. 
During this stage, we also gather BibTeX references for datasets to link and retrieve the original papers describing these datasets from arXiv.

\begin{table*}[ht]  
    \centering  
    \resizebox{\textwidth}{!}{
    \small 
    \setlength{\tabcolsep}{2pt} 
    \renewcommand{\arraystretch}{0.9} 
    \begin{tabular}{c|c|c|c|p{4cm}|c|c|c|c|c}  
    
    \toprule
    
    \textbf{ArXiv ID} & \textbf{Table} & \textbf{Dataset} & \textbf{Subset} & \textbf{\quad\quad
    Dataset Description} & \textbf{Model} & \textbf{Shots} & \textbf{Prompt} & \textbf{Metric} & \textbf{Value} \\ \midrule  
    2301.08721 & 2 & CommonsenseQA & xx &   

\textbf{Dataset Summary} This dataset is a multiple-choice question answering dataset focused on commonsense knowledge...
& GPT-4 & 12 & Batch CoT & Acc & 86 \\ \midrule

2408.02718 & 3 & MMIU & Overall &   
    
\textbf{Dataset Summary} This dataset is designed for the domain of multimodal understanding, specifically focusing on tasks that require the integration of information from multiple images...
& GPT-4o & xx & xx & Accuracy & 55.7 \\ \midrule

       2405.02861 & 3 & LexBench & IED &   
 \textbf{Dataset Summary} This dataset is part of a comprehensive evaluation suite designed to test language models on various semantic phrase processing tasks...
& Claude3 & 0 & zero-shot & Accuracy & 66.3 \\ \midrule

       2409.19667 & 5 & ProGraph & BGT &   

\textbf{Dataset Summary} This dataset is a benchmark designed to evaluate the capability of large language models (LLMs) in graph analysis tasks...
& Gemini1.0 & xx & xx & Accuracy & 27.7 \\

\bottomrule  
    \end{tabular}}  
    \caption{Sampled dataset instances. 
    ‘xx’ indicates missing values from the paper.
    Dataset descriptions, including summaries, tasks, and subsets, are abbreviated due to space limits. Full versions are in \cref{appendix:dataset_examples}.
    }
    \label{tab:main-dataset-instances}  
\end{table*}

\subsection{Dataset Description Generation}
\label{subsection:generation}

In addition to extracting structured representations of experiments, we also generate summary descriptions of the relevant tasks and datasets (Fig. \ref{fig:data_description}) that can aid in-depth literature analysis, such as classifying records by research subarea. 
We use a two-stage approach to create these descriptions. 
Initially, the LLM generates descriptions using its internal knowledge based on the dataset name and subset, which is cost-effective.
When the LLM is uncertain and refuses to answer, we prompt it to extract descriptions using the full content of the source papers as reference.
The source papers can be either the paper containing the table or the original dataset paper cited within it, which is retrieved using the BibTex references obtained from \cref{subsection:extraction}.

\section{Comprehensive Analysis of the Dataset}
\label{sec:statistics_and_quality}


\begin{table}[t!]
\centering
\resizebox{0.89\columnwidth}{!}{
\begin{tabular}{lc}
\toprule
\textbf{Metric} & \textbf{Value} \\
\toprule
\multicolumn{2}{l}{\textbf{General Statistics}} \\
Total Records & 18,127 \\
Number of Unique Datasets & 2,984 \\
Number of Table Source Papers & 1,737 \\
Number of Unique Tables & 2,694 \\
\hline
\multicolumn{2}{l}{\textbf{Records per Model}} \\
GPT-4~\citep{achiam2023gpt} & 12,475 \\
GPT-4o~\citep{hurst2024gpt} & 4,589 \\
Claude3-Opus~\citep{anthropic@claude} & 661 \\
Gemini1.0-Pro~\citep{team2023gemini} & 402 \\
\hline
\multicolumn{2}{l}{\textbf{Missing Values (`xx')}} \\
Records missing \textit{Subset} & 2,892 \\
Records missing \textit{Prompting Method} & 5,489 \\
Records missing \textit{Number of Few-Shots} & 9,081 \\
\hline
\multicolumn{2}{l}{\textbf{Dataset Description Sources}} \\
Source Paper of the Table & 9,334 \\
GPT-4o~\citep{hurst2024gpt} & 7,199 \\
Source Paper of the Linked Dataset & 1,594 \\
\bottomrule
\end{tabular}}
\caption{Dataset Statistics Overview. 
Dataset description sources specify whether descriptions come from LLM internal knowledge or source papers.}
\label{tab:dataset_stats}
\end{table}


Table \ref{tab:main-dataset-instances} presents sampled instances of \datasetname, while the dataset's statistics are summarized in Table \ref{tab:dataset_stats}.
A total of 18,127 experimental records of four target models were extracted from scanning over 300k arXiv source papers.
GPT-4~\citep{achiam2023gpt} and GPT-4o~\citep{achiam2023gpt} dominate the results, while Claude 3 Opus~\citep{anthropic@claude} and Gemini 1.0 Pro~\citep{team2023gemini} have significantly fewer entries, highlighting a bias favoring certain proprietary models. 
Many prompting configurations have missing values, confirmed by human verification~(\cref{subsec:quality_assessment}) as unreported by the original authors, likely indicating the default setup.
The majority of dataset descriptions are generated referencing the source paper of the table, as these papers often describe datasets when presenting benchmarks or detailing experimental sections.

\subsection{Quality and Efficiency Assessment}
\label{subsec:quality_assessment}

Our human evaluation assessed extraction accuracy and description quality across 40 records of \datasetname~(280 total attributes) from different papers. 
Two NLP Ph.D. students verified field extraction correctness~\citep{bai2023schema} and rated descriptions on a 5-point scale from (1) unrelated to (5) fully relevant and accurate~\citep{amidei2019use, liu2023geval}. 
Missing values were marked correct if the information was genuinely unavailable in the source papers.
More details are in \cref{appendix:human_evaluation}.

Table~\ref{tab:dataset_human} shows strong extraction accuracy and description quality of \datasetname. 
GPT-4o \citep{hurst2024gpt}, which we used during the data extraction, demonstrated effective long-context information extraction, showcasing its potential for scientific literature synthesis.
Moreover, high validation scores suggest missing values stem from unreported setups. 
Cohen's Kappa scores of 0.68 (extraction) and 0.57 (description) indicate substantial inter-annotator agreement~\citep{mchugh2012interrater}.

During the study, we recorded the time experts spent annotating target attribute information (excluding descriptions) from additional samples based on a list of papers and tables indexing experimental results of target models. 
On average, it took 7 minutes and 50 seconds per table, totaling approximately about 350 hours for \datasetname~of 2,694 tables. 
This does not include the initial effort of surveying and identifying papers and tables. 
In contrast, our pipeline identifies and extracts data from arXiv sources from 2023 to 2024 in a single day for under \$500 using a batching API.

\begin{table}[t!]
\centering
\resizebox{0.97\columnwidth}{!}{
\begin{tabular}{lc}
\toprule
\textbf{Attribute} & \textbf{Accuracy (Score)} \\
\toprule
\multicolumn{2}{l}{\textbf{Extraction}} \\
Dataset Name & 95 \% \\
Model Name & 100 \% \\
Prompting Method & 86.3 \% \\
Number of Few-Shot Examples & 95 \% \\
Metric & 100 \%  \\
Metric Value & 98.8 \% \\
\hline
\multicolumn{2}{l}{\textbf{Description}} \\
Dataset Description & 4.55 \\
\bottomrule
\end{tabular}}
\caption{Results of Human Evaluation.}
\label{tab:dataset_human}
\end{table}

\subsection{Categorization by Required Skills}
\label{subsection:core_skills}

We categorize experimental records from \datasetname~by required skills to enable comprehensive literature analyses and offer researchers a searchable resource. 
Expanding on Tulu3's core skills~\citep{lambert2024t}, we defined 10 categories: \textit{Knowledge}, \textit{Reasoning}, \textit{Math}, \textit{Coding}, \textit{Multimodality}, \textit{Instruction Following}, \textit{Safety}, \textit{Multilinguality}, \textit{Tool Use (Agent Framework)}, and \textit{Other}.
Records were classified using an LLM API~\citep{hurst2024gpt} based on dataset names, subsets, and descriptions. 
Since datasets can fit multiple categories, we applied multi-label classification. 
This categorized information is used throughout the analyses in \cref{section:literature_analysis_prompting_behavior}.

\begin{figure}[t!]
    \centering
    \includegraphics[width=0.85\columnwidth]{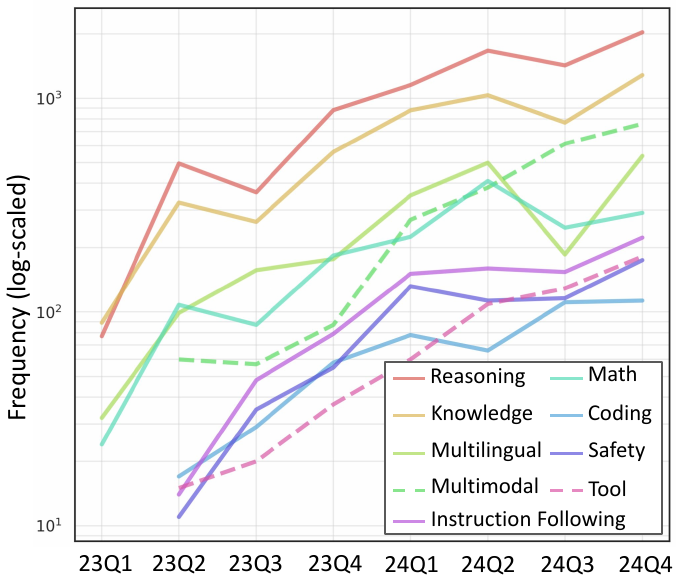}
    \caption{Log-scale frequency trends show a rapid increase of evaluation studies over successive quarters (Q). \emph{Reasoning} remains the most popular evaluation category, while \emph{Multimodality} exhibits recent rapid growth.}
    \label{fig:trend}
\end{figure}

Figure~\ref{fig:trend} shows the log-scale frequency of skill categories evaluated every three months based on arXiv publication dates. 
Over time, experimental results for all skills have increased due to rising interest in LLMs. 
\textit{Reasoning} tasks are the most popular and continue to grow, while \textit{Multimodality} has shown recent growth. \textit{Knowledge}, \textit{Multilinguality}, and \textit{Math} have steadily increased in evaluation rates. 
Frequency is based on the unique count of dataset-subsets aggregated by paper. 
The most frequently used datasets for each category are listed in \cref{appendix:frequent_datasets}.
\section{Prompting Behavior in Frontier LLMs: A Literature Review}
\label{section:literature_analysis_prompting_behavior}

\datasetname~enables partially automated literature analyses of LLM-prompting behaviors through its structured attributes.
We validate our semi-automated approach to literature analysis by replicating \citet{sprague2024cot}'s manual analysis, confirming CoT's advantages over direct prompting in mathematical and symbolic reasoning tasks (\cref{subsection:effect_cot}). 
We then show that \datasetname~enables a fine-grained analysis, showing that in-context examples boost performance in coding \& multimodal tasks but not in mathematical ones (\cref{subsection:effect_icl}). 
We qualitatively analyze dataset characteristics, such as required expert knowledge, that may negatively affect the performance of CoT and ICL (\cref{subsection:negative_analysis_dataset_characteristics}).
Moreover, we analyze CoT and ICL interactions, finding that while demonstrations enhance CoT performance, they do not affect CoT's relative improvement over standard direct prompting (\cref{subsection:joint_behavior}).
Finally, we analyze a subset of peer-reviewed papers published in major journals or conferences, revealing consistent trends that further support the robustness and generalizability of our findings (\cref{subsection:subset_analysis}).

\subsection{Which Tasks Benefit from CoT?}
\label{subsection:effect_cot}

\begin{figure}[t!]
    \centering  
    \includegraphics[width=0.9\columnwidth]{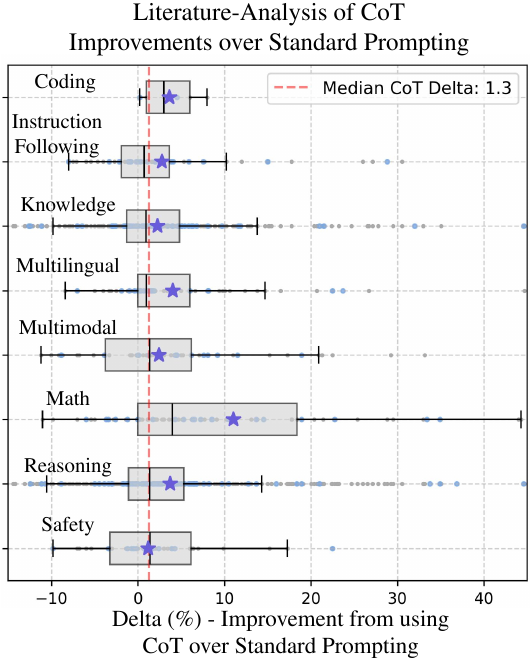}
    \caption{CoT shows significant performance improvement over direct prompting in mathematical tasks, whereas its impact on reasoning tasks is less distinct due to their complexity and diversity.
    Grey dots indicate individual deltas (improvements), blue dots represent the mean delta per paper, and a purple star marks the mean delta for each category.}
    \label{fig:cot_analysis_our_category}
\end{figure}

\paragraph{Motivation and Setup} 
CoT~\citep{wei2022chain}, a prompting technique for eliciting reasoning, has attracted considerable attention, with extensive literature on the subject \citep{wang2023towards, yao2024tree}
Recently, \citet{sprague2024cot} conducted a manual analysis, concluding that \textit{CoT improves over standard direct prompting primarily in mathematical and symbolic reasoning tasks}.  
We extract instances from \datasetname~that meet the criteria for replicating their investigation. 

This identification is achieved through filtering using the dataset's attributes. 
We focus on instances from the same source paper and table, identifying experiments that feature both CoT and standard direct prompts under the same conditions (model, dataset, subset, metric, and few-shot setup).
Human input is used only to identify whether a prompt is a CoT prompt or a standard direct prompt based on the \textit{prompting method} attribute; this is the only manual effort required for the analysis aside from implementing the analysis itself. 
We exclude CoT variations like "xx of Thoughts" \citep{yao2024tree} and CoT-SC \citep{wang2022self}. 
We then calculate CoT's performance improvement over standard direct prompting, namely \emph{delta}, where a positive delta indicates CoT outperforming direct prompting, and a negative delta indicates the reverse.


\begin{figure}[t!]
    \centering  
    \includegraphics[width=0.9\columnwidth]{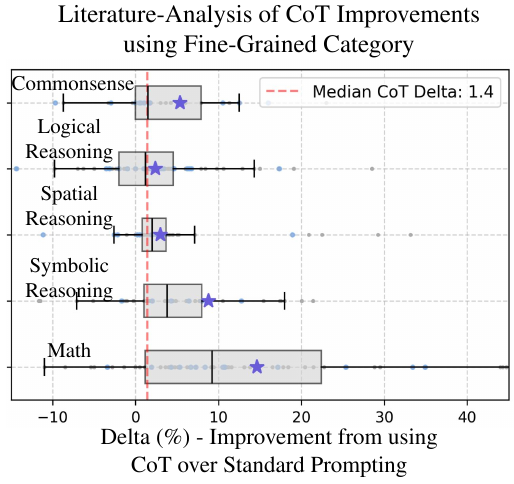}
    \caption{In the reasoning category, CoT shows notable improvement over direct prompting in \emph{Symbolic Reasoning} tasks compared to other reasoning tasks. The figure displays an abbreviated version of our study, using categories defined by \citet{sprague2024cot}.
    }
    \label{fig:cot_improvement_verification}
\end{figure}

\paragraph{Analysis} 
Our analysis, shown in Fig.\ref{fig:cot_analysis_our_category}, reveals that CoT significantly enhances performance in mathematical tasks, with both median and mean improvements surpassing the overall median. However, we do not see clear improvements for reasoning tasks, likely due to their diversity, such as commonsense and logical reasoning. 

To investigate further, we apply more detailed classification categories (but less general) from \citet{sprague2024cot} and redo our study using LLM classification. 
The truncated results, presented in Fig.~\ref{fig:cot_improvement_verification}, clearly show that symbolic reasoning and mathematical tasks exhibit notable improvement over other reasoning and non-reasoning tasks. 
This confirms the reliability and efficiency of \datasetname~for literature analyses, as it provides consistent results with those obtained through manual curation in previous studies. 
Statistical tests, as shown in \textit{Total Results} from Table. \ref{tab:statistical-tests}, confirm these patterns.

\subsection{Which Tasks Benefit from ICL?}
\label{subsection:effect_icl}

\paragraph{Motivation and Setup}

Previous research has systematically explored ICL behavior~\citep{min2022rethinking, agarwal2024many, wei2023larger, bertsch2024context}. 
In line with our study in~\cref{subsection:effect_cot}, we analyze the improvement of using in-context examples compared to not using any. 
To achieve this, we extract instances from \datasetname~that come from the same papers and tables comparing few-shot and zero-shot setups under identical conditions (i.e., model, dataset, subset, metric, and prompting method).
No manual labor was needed beyond implementation, as the \textit{number of in-context examples} is an integer value that can be easily computed for filtering. 
We measure performance deltas between few-shot and zero-shot setups, where positive values indicate few-shot advantages and negative values indicate the reverse.

\begin{figure}[t!]
    \centering
    \includegraphics[width=0.9\columnwidth]{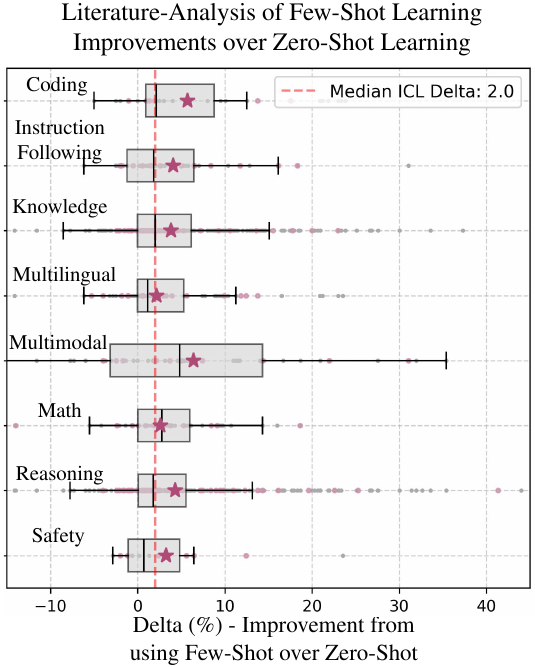}
    \caption{
    Few-shot learning significantly improves performance over zero-shot learning in \emph{coding} and \emph{multimodal} tasks, whereas its impact on math tasks is less pronounced compared to CoT, as shown in Fig.~\ref{fig:cot_analysis_our_category}.
     Grey dots represent individual deltas (improvements), sky pink dots show the mean delta per paper, and a pink star marks the mean delta for each category.}
    \label{fig:icl_improvement}
\end{figure}

\paragraph{Analysis}
Figure \ref{fig:icl_improvement} presents the results. 
In contrast to CoT's strong performance in mathematical reasoning, ICL shows only modest benefits for math tasks.
However, ICL demonstrates more substantial improvements in coding and multimodal applications despite considerable performance variability across different cases.
This variance may be due to the broad scope of multimodal tasks, including speech and image processing. 
ICL shows more uniform improvements over zero-shot performance across different categories compared to the study in Fig.~\ref{fig:cot_analysis_our_category} (std: 8.3).
Note that we did not account for the number of in-context examples used; we only distinguished between the use of in-context examples and zero-shot learning. 
We also plot the performance improvement distribution of more versus fewer in-context examples in \cref{appendix:details_in_meta_analysis} and find a similar overall median. 
This suggests that the presence of demonstrations is more important than their quantity.

\subsection{Which Dataset Characteristics Negatively Affect CoT and ICL?}
\label{subsection:negative_analysis_dataset_characteristics}

\begin{table}[t!]
\centering
\resizebox{0.9\columnwidth}{!}{
\begin{tabular}{l c c}
\toprule
\textbf{Key Characteristic} & \textbf{CoT Ratio} & \textbf{ICL Ratio} \\
\toprule
Expert Knowledge & 31.6\% & 31.0\% \\
Faithfulness & 20.9\% & 4.6\% \\
Complex Reasoning & 17.9\% & 13.8\% \\
Information Synthesis & 9.7\% & 10.3\% \\
Cognitive Tasks & 8.7\% & - \\
Affective Analysis & 7.1\% & 13.8\% \\
Structured Prediction & - & 12.6\% \\
Other & 2.5\% & 13.8\% \\
\bottomrule
\end{tabular}}
\caption{A qualitative analysis of dataset characteristics that resulted in decreased performance with CoT or ICL, compared to direct prompting or zero-shot set-up.
Key characteristics refer to the representative traits identified through qualitative analysis. The CoT and ICL ratios indicate the proportion of these traits among all negative cases for each method.}
\label{tab:dataset_negative_analysis}
\end{table}

\paragraph{Motivation and Setup} 
To comprehend the patterns of performance degradation, we analyze cases where performance declined: specifically, where CoT resulted in worse outcomes than direct prompting and where few-shot learning performed worse than zero-shot learning (i.e. deltas below zero). 
Table~\ref{tab:dataset_negative_analysis} summarizes the key dataset characteristics associated with performance declines under both approaches based on analyzing the descriptions.

\paragraph{Analysis} 
Out of the cases of performance decline, tasks requiring expert-level knowledge showed the highest ratio for both cases, approximately 31\%, indicating that knowledge-intensive tasks do not benefit much from different prompting configurations alone. 
\citet{sprague2024cot} notes a similar point that apparent performance gains in knowledge tasks mainly stem from "reasoning" or "math" components in datasets like MMLU~\citep{hendrycksmeasuring}.  
This also aligns partly with \citet{liu2024mind}, who found that CoT can hinder LLM performance on tasks requiring minimal deliberative thinking.  

The two approaches show different patterns in other aspects. 
CoT significantly degraded performance in faithfulness and factual verification tasks, possibly due to introduced reasoning biases \citep{shaikh2023second, chua2024bias}. 
In contrast, demonstrations led to notable declines in affective analysis and structured prediction tasks, where examples may bias the model toward imitating specific patterns rather than genuine reasoning. 
This also partially aligns with the finding from \citet{zhang2022robustness} that demonstrations increase the confidence of model predictions on captured superficial patterns, possibly incurring biases for structure prediction tasks such as NER.

\begin{table}[t]
\centering
\resizebox{\columnwidth}{!}{
\begin{tabular}{lccc}
\toprule
\textbf{Performance comparison} & \textbf{Median} & \textbf{Q1} & \textbf{Q3} \\ 
\midrule
Few-shot CoT $-$ Zero-shot CoT & 3.0 & 0.4 & 9.2 \\
\bottomrule
\end{tabular}}
\caption{Effect of in-context demonstrations \emph{within} Chain-of-Thought prompting.}
\label{tab:cot-demo-improvements}
\end{table}

\begin{table}[t]
\centering
\resizebox{\columnwidth}{!}{
\begin{tabular}{lccc}
\toprule
\textbf{Demonstration level} & \textbf{$\Delta$(Median)} & \textbf{$\Delta$(Q1)} & \textbf{$\Delta$(Q3)} \\
\midrule
Zero-shot           & +1.3 & $-$0.4 & 4.7 \\
Few-shot            & +0.9 & $-$1.2 & 3.7 \\
\bottomrule
\end{tabular}}
\caption{The performance comparison ($\Delta$) between CoT and standard prompting under matched demonstration settings. A positive $\Delta$ indicates that CoT outperforms standard prompting.}
\label{tab:cot-vs-standard}
\end{table}

\begin{table*}[ht]  
\centering  
\small  
\renewcommand{\arraystretch}{1.3}  
\begin{tabular}{l|c|cc|c|cc}  
\toprule  
\multirow{2}{*}{\textbf{Category}} & \multicolumn{3}{c|}{\textbf{Total Results}} & \multicolumn{3}{c}{\textbf{Filtered Results}} \\  
& Mean $\Delta$ & p-value & Significant & Mean $\Delta$ & p-value & Significant \\  
\midrule  
Math & 14.61 & 0.0000 & Yes & 13.53 & 0.0000 & Yes \\  
Symbolic and algorithmic & 8.85 & 0.0002 & Yes & 9.13 & 0.0000 & Yes \\  
Spatial and temporal reasoning & 3.03 & 0.0166 & No & 2.07 & 0.0056 & No \\  
Logical reasoning & 2.39 & 0.0084 & No & 1.18 & 0.3776 & No \\  
Commonsense reasoning & 5.41 & 0.0450 & No & 5.61 & 0.0748 & No \\  
Multi-hop QA & 2.05 & 0.0000 & Yes & 1.21 & 0.0064 & No \\  
Context-aware QA & 2.45 & 0.0014 & Yes & 0.28 & 0.7060 & No \\  
Encyclopedic knowledge & 2.18 & 0.0076 & No & 3.31 & 0.0138 & No \\  
Generation & 4.24 & 0.0280 & No & 1.92 & 0.3920 & No \\  
Text classification & 1.01 & 0.4464 & No & 0.27 & 0.8644 & No \\  
Entailment & 0.81 & 0.2070 & No & 0.93 & 0.1666 & No \\  
\bottomrule  
\end{tabular}  
\caption{Statistical test results across different categories for replicating the study from \citet{sprague2024cot}. \textbf{Total Results} refers to the results without filtering any papers. \textbf{Filtered Results} refer to the results from the filtered papers that are published in peer-reviewed journals \& conferences. Mean $\Delta$ shows the average improvement when using CoT over standard prompting, and significance is determined at $p=0.00227$ after applying a Bonferroni correction.}  
\label{tab:statistical-tests}
\end{table*}

\subsection{How do CoT and ICL Interact?}
\label{subsection:joint_behavior}



\paragraph{Motivation and Setup}
Building on previous work \cref{subsection:effect_cot,subsection:effect_icl}, we ask two related questions about the joint behavior of CoT and ICL:
\begin{enumerate}
    \item \textbf{Do demonstrations boost performance inside the CoT framework?}  
          We compare few-shot and zero-shot settings while keeping CoT prompting constant (Table~\ref{tab:cot-demo-improvements}), replicating the manual study from \citet{wei2022chain} through our literature analysis.
    \item \textbf{Do demonstrations enable CoT to beat standard prompting?}  
          We compare CoT and standard prompting under the same demonstration level (Table~\ref{tab:cot-vs-standard}).
\end{enumerate}

\paragraph{Analysis}
\begin{enumerate}
    \item \textbf{Demonstrations within CoT.}  
          Table~\ref{tab:cot-demo-improvements} shows that adding demonstrations yields a median gain of \textbf{+3.0 pts} which is significant.  
          This replicates the uplift reported by~\citep{wei2022chain, zhang2022automatic}.
    \item \textbf{Demonstrations do \emph{not} “rescue” CoT.}
        Table~\ref{tab:cot-vs-standard} shows that, whether we include demonstrations or not, CoT fails to open a meaningful lead over standard prompting.
        The median gains are modest— \textbf{+0.9 pts} in the few-shot setting and \textbf{+1.3 pts} in zero-shot.
        In other words, demonstrations lift both prompting styles in parallel but do not turn CoT into the clearly better option.
\end{enumerate}

\subsection{How Do Trends in Peer-Reviewed Papers Compare to Those in arXiv Publications?}
\label{subsection:subset_analysis}

\paragraph{Motivation and Setup}

\datasetname~includes information from papers published on arXiv. To examine whether the trends observed in our general analysis align with those in more selective venues, we conducted a focused analysis on a subset of peer-reviewed papers accepted at journals or conferences, using metadata from DBLP.\footnote{\url{https://dblp.org/}}
We report statistical significant test results from \ref{subsection:effect_cot} using a subset of peer-reviewed papers. Following \citet{sprague2024cot}, we perform a one-sided bootstrap test to assess whether each category shows performance gains (mean improvement > 0).

\paragraph{Analysis} 

Table~\ref{tab:statistical-tests} represents the results.
The statistically significant test results from our \textbf{Total Results} and \textbf{Filtered Results} indicate that math and symbolic reasoning tasks derive substantial benefit from using Chain-of-Thought (CoT) prompting, which aligns with both manual analysis~\citep{sprague2024cot} and our examination in \cref{subsection:effect_cot}.
We can see that the core finding remains consistent even if we used a subset of peer-reviewed papers.
This invariance in analysis patterns extends to other examinations like those in \cref{subsection:joint_behavior}, with additional details provided in Appendix~\ref{appendix:details_in_meta_analysis}.
These consistent results across both the complete dataset and the filtered selection of high-quality published papers serve to corroborate our findings and implicitly confirm our hypothesis.

\section{Related Work}

\paragraph{Information Extraction} Previous works have focused on extracting basic result tuples (e.g., task, dataset, metric) from scientific literature~\citep{singh2019automated, hou2019identification, kardas2020axcell, yang2022telin, bai2023schema, singh2024legobench, csahinucc2024efficient, kabongo2024orkg}. 
Our extraction pipeline improves upon this approach in two significant ways: it extracts enriched tuples that include prompting-related attributes and generates detailed dataset descriptions by leveraging LLM and automatically linked source papers.
Hence, unlike previous works that primarily compiled leaderboard tables, our enhanced extraction pipeline enables deeper review analysis, contributing to the broader goal of AI-driven scientific discovery~\citep{xu2021artificial, majumder2024discoverybench, m2024augmenting}.

\paragraph{LLM \& Prompting} Our study focuses on extracting experimental results of frontier proprietary LLMs~\citep{achiam2023gpt, anthropic@claude, team2023gemini}, with a specific emphasis on target attributes that incorporate information about prompting methods~\citep{brown2020language, wei2022chain}. 
In the context of prompting, prior studies have analyzed the mechanisms behind prompting methods, focusing either on the use of in-context examples~\citep{min2022rethinking,lampinen2022can,weber2023mind, zhang2022robustness} or techniques that elicit reasoning, such as CoT prompting~\citep{wei2022chain, shaikh2023second, wang2023towards, turpin2024language}.  
Conversely, we examine the model's behavior by conducting a literature analysis, which compiles data from scientific sources to reveal insights.

\paragraph{Literature Analysis}

Literature analysis systematically aggregates and examines data from multiple independent studies on a given topic to derive more precise and reliable conclusions. 
It has been widely applied in the biomedical domain for identifying target materials or clinical records~\citep{bao2019using, yun2024automatically}. 
In the NLP domain, review analysis has been used for metric standardization~\citep{reiter2018structured}, literature review~\citep{santu2024prompting, du2024llms}, and assessing evaluation criteria for specific domains~\citep{ostheimer2023call}. 
In contrast, our work employs a review analysis approach to evaluate the behavior of LLMs.
In the context of LLMs, \citet{asai2024openscholar} utilizes retrieval-augmented language models to synthesize scientific literature. 
However, this approach can only process a limited number of documents during retrieval to synthesize.
The work by~\citet{sprague2024cot} is perhaps the most closely related to ours. 
They conducted a review analysis through a literature survey to examine the effectiveness of CoT prompting for LLMs. 
However, their study is focused on CoT prompting, conducted on a limited scale, and relies on manual extraction methods.

\section{Conclusion}
\label{sec:conclusion}

Our study streamlines literature analysis by using an LLM for dataset extraction, demonstrating that the automatically extracted dataset, \datasetname, can yield novel findings and replicate manual analyses.
 We confirm the dataset's quality by replicating a key finding from \citet{sprague2024cot}. 
 Our analysis provides insights into prompting configurations, such as the benefits of ICL, the combined effects of CoT and ICL, and the dataset characteristics on performance declines with CoT or ICL.
 Overall, our resources support ongoing literature analyses, enhancing the understanding of LLM behaviors.
\section{Limitations}
\label{sec:limitation}

\paragraph{Target Model Scope}
Our study focuses on four leading proprietary LLMs selected for their widespread adoption and extensive documentation in existing literature. 
We excluded newer models like GPT4-o1 and Deepseek-R1 due to limited published results, though our analysis pipeline can easily accommodate them as more experimental data becomes available. 
While most analyzed models are not fine-tunable, this limitation presents an opportunity for future research combining fine-tuning and prompting methods analysis.

\paragraph{Further Validation}

Our literature analysis aims to identify trends by aggregating information across multiple studies, which generates potential hypotheses but requires systematic validation. 
While we report findings and patterns observed in the aggregated scientific literature, we did not independently validate each claim or finding. 
This limitation suggests the need for future work to rigorously test the hypotheses emerging from our analysis.

\paragraph{Attributes' Descriptiveness}
We frequently observed that the extracted attributes were not descriptive enough, which can hinder the dataset's utility for further analysis. 
Techniques like \textit{Batch COT}, for example, would benefit from more detailed descriptions. 
Additionally, the dataset descriptions could be enhanced to better differentiate between various dataset characteristics. 
Our attempts to further generate the ``collection process'' of the dataset resulted in numerous inaccuracies. Moreover, efforts to automatically link descriptions to actual dataset instances also encountered technical challenges, necessitating extensive manual intervention. 
Future work should aim to develop more effective methods for comprehensive dataset characterization.

\paragraph{Dataset Canonicalization}
Cross-study analysis requires standardizing dataset names and formats. 
While we implemented strict rules for dataset canonicalization, our approach likely missed potential matches. 
Alternative matching techniques we explored using LLM produced too many false negatives, whereas linking to the PaperswithCode dataset ontology\footnote{https://paperswithcode.com/datasets} was limited by its incomplete coverage of datasets.

\section*{Acknowledgments}
We would like to thank Microsoft's Azure Accelerate Foundation Models Research Program and NVIDIA's Academic Grant Program for providing computational resources to support this work.  This research is supported in part by the NSF under grant numbers IIS-2052498 and SMA-2418946. Any opinions, findings, and conclusions or recommendations expressed in this material are those of the author(s) and do not necessarily reflect the views of the National Science Foundation.
We also appreciate Ethan Mendes, Duong Minh Le, and Seongeun Park for their valuable discussions and feedback on the paper.

\bibliography{acl_latex}

\appendix
\clearpage

\section{Details and Trials During Extraction}
\label{appendix:data_extraction}

We provide the details and trials during the data extraction process.

\paragraph{Hyperparameters and Prompt Tuning}

We did not extensively tune any LLM hyperparameters, instead, we used OpenAI’s default generation setup with greedy decoding.
For prompting, we initially built upon previous work~\citep{bai2023schema} as a foundation. 
Still, we adapted the prompts to better suit our project’s objectives, such as (1) determining whether the table includes experimental results for our target model or (2) enhancing records using contextual information from the paper.

\paragraph{Preprocessing} To extract the experimental results of target models and attributes, we utilize arXiv sources\footnote{\url{https://info.arxiv.org/help/bulk_data_s3.html}} published between January 2023 and December 2024, as this timeframe aligns with the release of proprietary models, ensuring relevance to the latest advancements in the field. 
We specifically targeted papers in the machine learning (ML) domain (cs.CV, cs.AI, cs.CL, cs.LG) and downloaded their LaTeX source files for detailed analysis. 
To extract structured data, such as tables, we utilized a regex-based rule extraction method to retrieve the LaTeX source of tables along with their indices~\citep{bai2023schema}.

\paragraph{Filtering}

Before the extraction stage, we filter tables to reduce extraction time and API costs. 
To cut computational expenses, we pre-filter tables for those containing results of target models. 
Using simple heuristics, we filter tables based on keywords related to the models (e.g., "gemini," "gpt"), which significantly reduces irrelevant data and minimizes LLM API usage.
Moreover, we focus on leaderboard tables, which present the main results on specific benchmarks~\citep{kardas2020axcell}, to assess LLM performance. 
We employ an LLAMA3.1-70B-Instruct~\citep{dubey2024llama} to filter out non-leaderboard tables.

\paragraph{Extraction and Augmentation}

We employ schema-driven extraction to obtain records matching our target attributes from tables (detailed in \Cref{subsection:attributes}). 
The LLM first determines if a table contains target model records and only proceeds with extraction when relevant models are present, skipping tables without target model data.
While our approach resembles \citet{bai2023schema}, it achieves significant cost savings while maintaining accuracy (described in \cref{subsec:quality_assessment}) by focusing solely on target model records rather than processing entire tables. 
This selective approach reduces API calls by up to 1/200 per table.

After extracting information from the table, we further augment the extracted records by incorporating additional context from the entire paper. 
This is important because valuable details, such as experimental specifics, are often found in various sections of the paper. 
We heuristically filter out irrelevant sections from the full latex source and provide the LLM with the extracted information to enhance the attribute details using the context. 
During this process, we also gathered BibTeX references for the datasets associated with each record, allowing us to link to and retrieve the original papers describing these datasets.

\paragraph{Dataset Description Generation}

Initially, we asked the LLM to generate descriptions using only the dataset name and subset information from its internal knowledge. 
This approach was cost-effective as it didn't require processing additional context. 
However, the LLM may be uncertain—particularly with lesser-known datasets or those beyond its knowledge cutoff. To address this, we instructed the model to refuse to answer when unsure about dataset information~\citep{bai2022constitutional}. 
In such cases, a second stage is triggered, where the model is prompted to extract the necessary information directly from the source papers.
When LLM knowledge was insufficient, we directly extracted dataset descriptions from source papers. 
These source papers could be either the paper containing the table being processed or the original dataset paper cited by it. 
We used citation tags from \cref{subsection:extraction} and rule-based heuristics to link to external arXiv papers associated with the original dataset.
We then used the full content of these source papers to prompt the LLM to extract the dataset descriptions again, using the paper’s contents as a reference.

We combined the dataset name and subset name to facilitate information generation or extraction. 
The schema for generation includes \textit{Dataset Summary}, \textit{Task Explanation}, and \textit{Subset Description}. 
Initially, we attempted to generate a \textit{Collection Process} section, detailing how the dataset was sourced and curated, but this led to excessive inaccuracies, so we decided to omit this part.

\paragraph{Models} We utilize GPT-4o \citep{hurst2024gpt} as the LLM for the pipeline and employ Llama-3.1 70B \citep{dubey2024llama} for filtering tables.

\paragraph{Valid Metrics} 

To ensure consistent analysis, we established a standardized set of metrics and excluded records using metrics outside this set. 
We also normalized all metric values to maintain uniformity across the analysis.
The approved metrics are: Accuracy, Exact Match, F1, BLEU, Rouge, MRR (Mean Reciprocal Rank), Precision, Recall, Pearson Correlation Coefficient, MAE (Mean Absolute Error), and MSE (Mean Square Error). 
For all metrics except Pearson correlation coefficient, MAE, and MSE, we standardized the values to range from a minimum of 0 to a maximum of 100 (e.g., 0.63\% was scaled to 63).

\paragraph{Missing Values} If certain target attributes were unavailable from the paper or the LLM couldn't locate specific information, we instructed it to mark these missing values with "xx" as a placeholder.

\paragraph{Canonicalization} 

Using rule-based heuristics, we canonicalized the \textit{Dataset name} by grouping identical names or methods under different labels. 
We took care to avoid merging similar but distinct names representing different versions, variations, or separate entities by applying strict character and abbreviation matching rules.
We initially explored linking datasets to the PaperswithCode ontology\footnote{\url{https://paperswithcode.com/datasets}}. 
However, the outdated and incomplete nature of its dataset list made this approach infeasible. 
We then experimented with clustering algorithms and large language models (LLMs) for grouping, but these methods resulted in an excessive number of false positives. 
Ultimately, we adopted a rule-based grouping algorithm as our approach.

\paragraph{Postprocessing} 

To ensure dataset quality, we removed duplicate records with identical values for \textit{Dataset Name}, \textit{Subset}, \textit{Number of Few\-Shots}, \textit{Prompting Method}, \textit{Metric Name} but differing \textit{Performance}. 
Additionally, we filtered out records with invalid dataset descriptions.


\section{Details in Human Evaluation}
\label{appendix:human_evaluation}

We detail the human evaluation process.

\paragraph{Extraction Evaluation}
The objective is to assess whether the fields within records have been accurately extracted. 
This evaluation involves checking each field in a record against the original source paper to confirm its correctness. 
Fields are marked with 'o' for success or 'x' for failure, depending on the accuracy of the extraction. 
Annotators were instructed to use the \textit{original\_extracted\_dictionary} for verification, as metric names have been standardized and canonicalized. 
A table index was provided to help locate records and tables more quickly for annotation. 
Additionally, it is permissible for prompting methods to include few-shot examples if relevant information is not found in the paper.

\paragraph{Description Evaluation}
This step evaluates whether the description aligns appropriately with the dataset-subset pair. 
The evaluation protocol (5-Point Likert Scale) involves assessing the quality of the dataset description based on the following rubric:  

\begin{itemize}
\item Score 1: The description is completely unrelated to the dataset-subset pair.  
\item Score 2: The description has minimal relevance but lacks alignment or context. 
\item Score 3: The description is moderately relevant, capturing the essence of the dataset-subset pair but includes noticeable inaccuracies.  
\item Score 4: The description is highly relevant with only minor inaccuracies.  
\item Score 5: The description is fully relevant and entirely accurate, with no errors.  
\end{itemize}

When scoring, annotators used references from the web or literature searches to ensure the evaluation was well-informed.
Our scoring rubric is designed to approximate an interval scale, allowing us to compute average scores. 
This approach aligns with standard practices in the machine learning field for evaluating response quality, whether through model-based or human-based assessments~\citep{liu2023g, kim2023prometheus}.
\section{Details in Literature Analysis of Prompting Behavior in Frontier LLMs}
\label{appendix:details_in_meta_analysis}

\subsection{Which Categories  Benefit from CoT?}

To replicate the Chain-of-Thought (CoT) improvements over standard direct prompting, as demonstrated by \citet{sprague2024cot}, we adopt their category schemas, which include entailment, text classification, generation, knowledge, contextQA, multihopQA, commonsense reasoning, logical reasoning, spatial reasoning, and math. 
While these categories provide a fine-grained definition of traditional NLP tasks, they lack generalizability in encompassing the broader capabilities of modern LLMs, such as multimodality, safety, and tool use. 
The complete version of Fig.~\ref{fig:cot_improvement_verification} is presented below in Fig.~\ref{fig:cot_improvement_verification_full}.

Extracting records for this type of analysis is challenging due to the strict criteria for sample selection. 
While \citet{sprague2024cot}'s dataset contains a larger overall sample size, as it is not limited to a single model and \datasetname~is not specifically oriented to replicating their study. 
For the overlapping model (GPT-4), their dataset includes 168 instances, whereas ours contains 553. This demonstrates the scalability of our pipeline.

\begin{figure}[t!]
    \centering  
    \includegraphics[width=1.0\columnwidth]{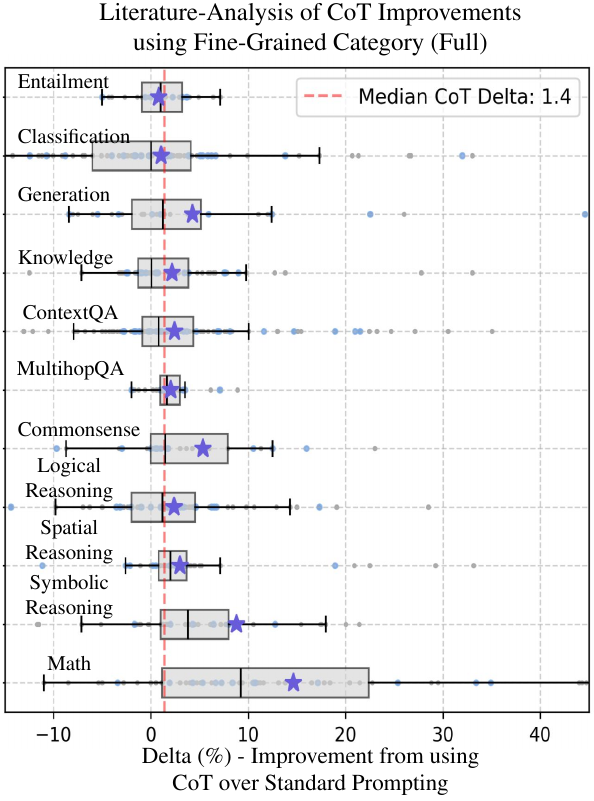}
    \caption{Full version of box plots showing performance improvements with CoT compared to standard prompting, categorized according to \citet{sprague2024cot} for a fine-grained investigation in various reasoning tasks.}
    \label{fig:cot_improvement_verification_full}
\end{figure}

\begin{table*}[t]
\centering
\resizebox{0.7\textwidth}{!}{
\begin{tabular}{lcccccc}
\toprule
 & \multicolumn{3}{c}{\textbf{Original}} & \multicolumn{3}{c}{\textbf{Filtered}} \\
\cmidrule(lr){2-4}\cmidrule(lr){5-7}
\textbf{Performance comparison} & Median & Q1 & Q3 & Median & Q1 & Q3 \\
\midrule
Few-shot CoT $-$ Zero-shot CoT & 3.0 & 0.4 & 9.2 & 5.0 & 2.3 & 10.4 \\
\bottomrule
\end{tabular}}
\caption{Effect of in-context demonstrations \emph{within} Chain-of-Thought prompting. \textbf{Original} uses all papers; \textbf{Filtered} keeps only top-venue papers.}
\label{tab:cot-vs-zero-shot-cot-subset}
\end{table*}

\begin{table*}[t]
\centering
\resizebox{0.6\textwidth}{!}{
\begin{tabular}{lcccccc}
\toprule
\textbf{Demonstration level} & \multicolumn{3}{c}{\textbf{Original }$\Delta$} & \multicolumn{3}{c}{\textbf{Filtered }$\Delta$} \\
\cmidrule(lr){2-4}\cmidrule(lr){5-7}
 & Median & Q1 & Q3 & Median & Q1 & Q3 \\
\midrule
Zero-shot & +1.3 & $-$0.4 & +4.7 & +1.3 & 0.0 & +3.6 \\
Few-shot  & +0.9 & $-$1.2 & +3.7 & +1.3 & 0.0 & +3.0 \\
\bottomrule
\end{tabular}}
\caption{CoT versus standard prompting when the demonstration setting is fixed. Positive $\Delta$ means CoT outperforms Direct prompting.\textbf{Original $\Delta$} uses all papers; \textbf{Filtered $\Delta$} keeps only top-venue papers.}
\label{tab:cot-vs-standard-subset}
\end{table*}

\begin{figure}[t!]
    \centering
    \includegraphics[width=0.9\columnwidth]{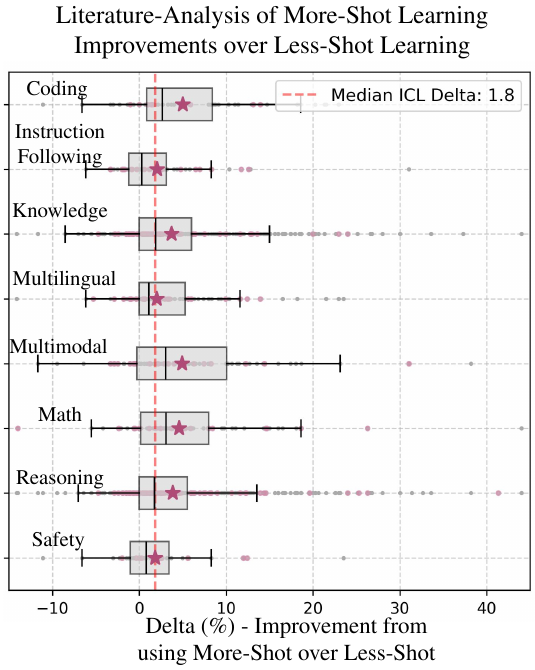}
    \caption{Box plot distributions showcasing performance enhancements using more in-context examples compared to less in-context examples setup, categorized by skill sets in \cref{subsection:core_skills}. 
     Grey dots represent individual deltas, while sky pink dots show the mean delta aggregated for each paper. A pink star indicates the mean delta for each category.}
    \label{fig:icl_improvement_more_vs_less}
\end{figure}

\subsection{Which Categories Benefit from ICL?}

We visualize the distribution of in-context examples, comparing cases with more versus fewer demonstrations, in Fig.~\ref{fig:icl_improvement_more_vs_less}. 
The results show that the overall median and distribution are similar to those in Fig.~\ref{fig:icl_improvement}, suggesting that the presence of demonstrations is more crucial than their quantity.

\subsection{How Do Trends in Peer-Reviewed Papers Compare to Those in arXiv Publications?}

We reanalyzed the joint behavior findings from \cref{subsection:joint_behavior} using only peer-reviewed papers published in journals or conferences.
Table~\ref{tab:cot-vs-zero-shot-cot-subset} and Table~\ref{tab:cot-vs-standard-subset} present these results.
The filtered data confirms our original finding regarding the interaction between CoT and ICL: CoT with demonstrations (ICL) consistently outperforms CoT without demonstrations. Additionally, CoT's improvement over standard prompting remains consistent regardless of demonstration count (zero-shot or few-shot).
\section{Dataset Examples}
\label{appendix:dataset_examples}

We provide a curated selection of examples from \datasetname~in Table~\ref{tab:dataset-instances} and Table~\ref{tab:dataset-instances-non-gpt}.

\begin{table*}[ht]  
    \centering  
    \resizebox{\textwidth}{!}{
    \small 
    \setlength{\tabcolsep}{2pt} 
    \renewcommand{\arraystretch}{0.9} 
    \begin{tabular}{c|c|c|c|p{8.5cm}|c|c|c|c|c}  
    
    \toprule
    
    \textbf{ArXiv ID} & \textbf{Table} & \textbf{Dataset} & \textbf{Subset} & \textbf{\quad\quad\quad\quad\quad\quad\quad\quad\quad
    Dataset Description} & \textbf{Model} & \textbf{Shots} & \textbf{Prompt} & \textbf{Metric} & \textbf{Value} \\ \midrule  
    2301.08721 & 2 & CommonsenseQA & xx &   

\textbf{Dataset Summary}: This dataset is a multiple-choice question answering dataset focused on commonsense knowledge. It is designed to evaluate a model's ability to understand and apply commonsense reasoning to answer questions correctly. The dataset is widely used in the natural language processing domain to benchmark the performance of models on tasks requiring commonsense understanding.

\textbf{Task Explanation}: The task involves providing a question along with multiple answer choices, where the input is the question and the list of possible answers. The output is the correct answer choice. The task is evaluated based on the accuracy of the model's predictions compared to the correct answers provided in the dataset.
& GPT-4 & 12 & Batch CoT & Acc & 86 \\ \midrule  

    2301.08745 & 8 & Flores-101 & De $\rightarrow$ En &   
    
\textbf{Dataset Summary}: This dataset is a multilingual translation benchmark designed to evaluate machine translation systems across a wide range of languages. It covers 101 languages, providing a comprehensive resource for assessing translation quality and performance. The dataset is intended for tasks involving language translation, with a focus on both high-resource and low-resource languages. Key characteristics include its diverse language pairings and standardized evaluation metrics, which aim to facilitate consistent and fair comparisons of translation models.

\textbf{Task Explanation}: The primary task associated with this dataset is machine translation. The input consists of text in one language, and the output is the translated text in another language. For this specific subset, the task involves translating text from German (De) to English (En). Evaluation of the task is typically conducted using metrics such as BLEU, which measures the accuracy and fluency of the translated output compared to reference translations.

\textbf{Subset Description}: The De$\rightarrow$En subset specifically focuses on translations from German to English. This subset is used to evaluate the performance of translation models in converting German text into English, providing insights into the model's ability to handle this particular language pair.
     & GPT-4 & 0 & Direct & BLEU & 46 \\ \midrule

       2303.07992 & 3 & KQApro & xx &   
    
\textbf{Dataset Summary}: This dataset is a large-scale knowledge-based question answering dataset designed to evaluate the ability of models to understand and reason over complex questions. It is primarily used in the domain of natural language processing and artificial intelligence, focusing on tasks that require comprehension of structured knowledge bases. The dataset features a diverse set of questions that test various reasoning skills, such as logical reasoning, comparison, and temporal reasoning. The evaluation goals are to assess the accuracy and reasoning capabilities of models in answering these questions correctly.

\textbf{Task Explanation} The primary task associated with this dataset is knowledge-based question answering. The input consists of a natural language question, and the output is the correct answer derived from a structured knowledge base. Models are expected to interpret the question, retrieve relevant information from the knowledge base, and perform the necessary reasoning to arrive at the correct answer. The task evaluation method typically involves measuring the accuracy of the model's answers compared to a set of ground truth answers. & GPT-4 & 0 & xx & Accuracy & 57.2 \\ \midrule

       2304.08244 & 3 & API-Bank & Call &   

\textbf{Dataset Summary}: This dataset is a comprehensive benchmark designed to evaluate and enhance the capabilities of Large Language Models (LLMs) in utilizing external API tools. It focuses on tool-augmented LLMs, assessing their abilities in planning, retrieving, and calling APIs. The dataset includes a wide range of domains and APIs, aiming to provide a realistic and diverse evaluation environment. The primary goal is to understand the effectiveness of current LLMs in tool usage, improve their capabilities, and identify challenges in leveraging tools.

\textbf{Task Explanation}: The task involves evaluating LLMs on their ability to interact with APIs based on user queries. The input consists of user queries and API documentation, while the output is the correct API call and response. The task is evaluated based on the correctness of API calls and the quality of responses, using metrics like accuracy and ROUGE-L.

\textbf{Subset Description}: The ""Call"" subset specifically focuses on evaluating the ability of LLMs to make API calls based on given queries when the APIs are known. This subset tests the basic capability of LLMs to interact with APIs directly without the need for retrieval or planning. & GPT-4 & 0 & Zero-shot & Rouge & 36.91 \\ \midrule

       2310.05915 & 1 & HotpotQA & xx &   

\textbf{Dataset Summary}: This dataset is a large-scale, high-quality question answering dataset designed to facilitate research in the domain of natural language processing, specifically in multi-hop question answering tasks. It contains questions that require reasoning over multiple documents to arrive at the correct answer. The dataset is intended to evaluate the ability of models to perform complex reasoning and synthesis of information across different contexts.

\textbf{Task Explanation}: The primary task involves providing a natural language question as input and expecting a concise answer as output. The questions are designed to require multi-hop reasoning, meaning that the answer cannot be found in a single document but requires synthesizing information from multiple sources. The task is evaluated based on the accuracy of the answers provided by the model, often using metrics such as exact match and F1 score to assess performance. & GPT-4 & xx & IO & Exact Match & 37.2 \\

\bottomrule  
    \end{tabular}}  
    \caption{Curated list of examples from \datasetname~(Target Model: GPT-4~\citep{achiam2023gpt}).}
    \label{tab:dataset-instances}  
\end{table*}

\begin{table*}[ht]  
    \centering  
    \resizebox{\textwidth}{!}{
    \small 
    \setlength{\tabcolsep}{2pt} 
    \renewcommand{\arraystretch}{0.9} 
    \begin{tabular}{c|c|c|c|p{9cm}|c|c|c|c|c}  
    
    \toprule
    
    \textbf{ArXiv ID} & \textbf{Table} & \textbf{Dataset} & \textbf{Subset} & \textbf{\quad\quad\quad\quad\quad\quad\quad\quad\quad Dataset Description} & \textbf{Model} & \textbf{Shots} & \textbf{Prompt} & \textbf{Metric} & \textbf{Value} \\ \midrule  
    
    2407.19619 & 1 & HPC & Fortran2C++ &   

\textbf{Dataset Summary}: This dataset is designed for the domain of high-performance computing (HPC) code translation, specifically focusing on translating between OpenMP Fortran and C++ code. It aims to train and evaluate large language models (LLMs) to enhance their translation capabilities between these two languages. The dataset is characterized by its diverse and representative collection of open-source OpenMP benchmarks, refined through code similarity tests. The evaluation goals include improving translation accuracy, as measured by CodeBLEU scores, and achieving human-level translation proficiency.

\textbf{Task Explanation}: The primary task is to translate code from OpenMP Fortran to C++. The input is a Fortran code snippet, and the output is its equivalent C++ translation. The task is evaluated using CodeBLEU, a metric that assesses the quality of code translation by considering both syntactic and semantic elements. Human evaluation is also employed to ensure the translations are correct, readable, and maintain the original code's functionality.

\textbf{Subset Description}: The Fortran to C++ translation subset focuses specifically on translating code from Fortran to C++. It serves the purpose of training models to understand and convert Fortran code into its C++ equivalent, addressing the challenges of translating between these two prominent HPC languages.
& GPT-4o & 0 & Zero-Shot & BLEU & 37.1 \\ \midrule  

    2408.02718 & 3 & MMIU & Overall &   
    
\textbf{Dataset Summary}: This dataset is designed for the domain of multimodal understanding, specifically focusing on tasks that require the integration of information from multiple images. It aims to facilitate research in areas such as image comparison, visual storytelling, and cross-image reasoning. The key characteristics of the dataset include a diverse set of image pairs or groups, each accompanied by annotations or questions that require understanding the relationships or narratives across the images. The evaluation goals are to assess the ability of models to comprehend and reason about multiple images simultaneously.

\textbf{Task Explanation}: The primary task involves taking multiple images as input and producing an output that demonstrates understanding of the relationships or narratives between them. This could involve answering questions, generating descriptive text, or identifying similarities and differences. The task evaluation method typically involves comparing the model's output to human-generated annotations or answers, using metrics such as accuracy, BLEU score, or human judgment for qualitative assessments.

\textbf{Subset Description}: The Overall subset encompasses the entire dataset, providing a comprehensive collection of all image pairs or groups and their associated annotations. This subset is intended for general evaluation and benchmarking of models on the full range of tasks supported by the dataset.
     & GPT-4o & xx & xx & Accuracy & 55.7 \\ \midrule

       2405.02861 & 3 & LexBench & IED &   
 \textbf{Dataset Summary}: This dataset is part of a comprehensive evaluation suite designed to test language models on various semantic phrase processing tasks. It focuses on idiomatic expression detection, which is one of the ten tasks included in the suite. The dataset aims to evaluate the ability of language models to understand and process idiomatic expressions, which are non-compositional phrases whose meanings cannot be deduced from their individual components. The evaluation goals include assessing model performance in classification, extraction, and interpretation tasks related to idiomatic expressions.

\textbf{Task Explanation}: The task involves detecting idiomatic expressions within a given context. The input consists of a sentence containing an idiomatic expression, and the output is a choice from multiple options that best describes the idiomatic expression's meaning. The task is evaluated using exact match accuracy, where the model's prediction is compared against the correct option.

\textbf{Subset Description}: The Idiomatic Expression Detection (IED) subset specifically focuses on identifying idiomatic expressions within sentences and selecting the correct interpretation from multiple options. This subset is designed to challenge language models in understanding the non-compositional nature of idiomatic expressions.
& Claude3 & 0 & zero-shot & Accuracy & 66.3 \\ \midrule

       2409.19667 & 5 & ProGraph & BGT &   

\textbf{Dataset Summary}: This dataset is a benchmark designed to evaluate the capability of large language models (LLMs) in graph analysis tasks. It focuses on enabling LLMs to process and analyze graphs using external APIs, similar to how human experts would approach such tasks. The dataset is crafted to assess the models' ability to generate code solutions for graph-related problems, rather than relying solely on reasoning over raw inputs. The benchmark includes 512 problems across three categories: basic graph theory, graph statistical learning, and graph embedding. The evaluation aims to measure the models' performance in leveraging programming libraries to solve graph tasks.

\textbf{Task Explanation}: The primary task involves generating Python code to solve graph-related problems using specified APIs. The input consists of a problem statement describing a graph task, and the output is the corresponding Python code that utilizes relevant APIs to solve the task. The evaluation method involves assessing the pass rate (the ratio of executable code) and accuracy (the ratio of correct answers from the executable code) of the generated solutions.

\textbf{Subset Description}: The Basic Graph Theory (BGT) subset focuses on fundamental concepts of graph theory, including types, properties, classical algorithms, and basic operations. Tasks in this subset may involve checking graph acyclicity, computing node centrality, or finding maximum cardinality matching.
& Gemini1.0 & xx & xx & Accuracy & 27.7 \\ 

\bottomrule  
    \end{tabular}}  
    \caption{Curated list of examples from \datasetname~(Target Models: GPT-4o~\citep{hurst2024gpt}, Claude3 Opus~\citep{anthropic@claude}, and Gemini1.0 Pro~\citep{team2023gemini}).}  
    \label{tab:dataset-instances-non-gpt}  
\end{table*}  
\section{Frequent Datasets by Skill Category}
\label{appendix:frequent_datasets}

We present a list of the top 10 most frequently used datasets in Table~\ref{tab:frequent_datasets}, aggregated according to our defined skill categories, in ~\cref{subsection:core_skills}. 
Frequency is measured by the number of unique papers that evaluate a dataset, without counting multiple experiments within the same paper. 
Some datasets appear in multiple categories due to multi-label categorization and the use of different subsets focusing on distinct skill areas.

\begin{table*}[ht]  
    \centering  
    \small
    \renewcommand{\arraystretch}{1.3} 
    \begin{tabular}{p{0.15\textwidth}|p{0.8\textwidth}}
    \toprule
    \textbf{Skill Category} & \textbf{Dataset} \\ \midrule  
    Reasoning & MMLU, CommonsenseQA, StrategyQA, HotPotQA, ARC Challenge,
    BIG-Bench Hard, Spider, DROP, Winogrande, ScienceQA \\
    \hline
    Knowledge & MMLU, MedQA, MedMCQA, PubmedQA, NQ, TruthfulQA, United States USMLE, TriviaQA, ScienceQA, Financial Phrase Bank \\
    \hline 
    Multilinguality & C-Eval, WMT22, FLORES-200, XQuAD, Flores-101, MKQA, Flores, MMMLU, SIB-200, TOEFL11 \\
    \hline 
    Multimodality & ScienceQA, EgoSchema, MMMU, DAIC-WOZ, MathVerse, MathVista, MMBench, ChartQA, Text-VQA, DocVQA \\
    \hline 
    Math & GSM8K, MATH, SVAMP, AQuA, AddSub, Multi-Arith, ASDIV, Multilingual Grade School Math, Single Eq, MathVista \\
    \hline 
    Coding & Spider, BIRD, Human Eval, MBPP, APPS, CodeReviewer, Exception Handling Code Generation, BigCloneBench, WikiSQL, SecurityEval \\
    \hline 
    Instruction & Mind2Web, SCAN, PromptCBLUE, InstructExcel, Len, RuLES, SFRES, SemEval, InstructDial++, Alpaca Eval \\
    \hline 
    Safety & AdvBench, ETHICS, Bias Benchmark for Question Answering (BBQ), HHH Alignment, SimpleSafetyTests, i2b2/UTHealth de-identification challenge dataset, Fifty Shades of Bias, LLM-generated emails, Tensor Trust, MULTITuDE benchmark \\
    \hline 
    Tool Use & API-Bank, NERetrieve, PubMed Retrieval and Synthesis (PubMedRS-200), student help requests, PowerPoint Task Completion (PPTC), ToolTalk, TaskBench, AgentBench-WB, Compliance Checking, CardBench \\
    \bottomrule
    \end{tabular}
    \caption{Top 10 Most Frequent Datasets per Skill Category. Frequency is measured by the number of unique papers evaluating a dataset, without counting multiple experiments within the same paper.}  
    \label{tab:frequent_datasets}  
\end{table*}  
\section{Prompts Used during Data Extraction}
\label{appendix:prompt_examples}

We provide the prompts used during the data extraction in Table~\ref{tab:prompt_example} and Table~\ref{tab:prompt_example_context_augmentation}.

\begin{table*}[ht]  
    \centering  
    \small
    \renewcommand{\arraystretch}{1.3} 
    \begin{tabular}{p{0.15\textwidth}|p{0.8\textwidth}}
    \toprule
    \textbf{Component} & \textbf{Prompt} \\ \midrule  
    Identifying Leaderboard Tables & \texttt {Determine if the given Table LaTeX represents a leaderboard table.}
    
\texttt {Leaderboard tables showcase the main results of the paper on a specific benchmark, often comparing these results with those from other studies.}

\texttt {Tables are NOT considered leaderboard tables if they focus on ablation studies, hyperparameter tuning, dataset statistics, or other supplementary experiments.}

\texttt {Respond with 'true' if the table is a leaderboard table, and 'false' otherwise, with no additional explanation.}

\textbf
{Input:}  

\texttt {Table LaTeX: }  

\textbf
{Output:}  

\texttt
{Classification Output: true/false}  \\ \hline

Schema-Driven Extraction & 

\texttt {Your task is to extract all numeric cells representing the experimental results of a specified target model from a table LaTeX source, following the provided template. When extracting results for the target model, exclude results from its variant models.} \newline

\texttt{For example:}

\texttt{- For GPT-4, exclude GPT-4o, GPT-4-v, or Deplot + GPT-4.}

\texttt{- For Claude3 Opus, exclude Claude3 Sonnet, Claude3 Haiku, Claude2, Claude 3.5.}

\texttt{- For Gemini 1.0 Pro, exclude Gemini 1.5, Gemini 1.5 Pro, Gemini Ultra, and Gemini Flash.}

\texttt{- For GPT-4o, exclude GPT-4, GPT-4-o1, GPT4-Turbo, and GPT4-V.} \newline

\texttt{However, include results from different versions of the target model across time periods (e.g., GPT-4, GPT-4-0828, GPT-4-0623, GPT-4-0314). If the table contains results where the target model is used for generation or evaluation, exclude those results. The goal is to extract results about the target model itself, not those where it is used as a tool. If no numeric cells related to the target model are found, output "<FAILED>".}

\texttt{During output, return only the extracted results in the following template. Do not provide explanations. For any unanswerable attributes, leave their value as "xx".}

\texttt{Template:
\{"value": "xx", "dataset": "xx", "dataset\_citation\_tag": "xx", "subset": "xx", "model\_name": "xx", "metric": "xx", "prompting\_method": "xx", "number\_of\_shots": "xx"\}}

\texttt{Field Descriptions:}

\texttt{- value: Extracted numeric cell value.}

\texttt{- dataset: Name of the dataset or benchmark (must be a proper noun, e.g., "synthetic dataset" is not acceptable).}

\texttt{- dataset\_citation\_tag: Citation tag for the dataset.}

\texttt{- subset: Dataset subset used (e.g., subtask, domain, split).}

\texttt{- model\_name: Name of the model used in the experiment.}

\texttt{- metric: Evaluation metric used.}

\texttt{- prompting\_method: Prompting method used (do not include shot count here).}

\texttt{- number\_of\_shots: Integer value representing the number of shots used.}

\textbf{Input:}

\texttt{Target Model: }

\texttt{Table LaTeX Source: }

\textbf{Output}
\texttt{Extracted Results:} \\ 

    \bottomrule
    \end{tabular}
    \caption{Prompts for identifying leaderboard tables and schema-driven data extraction, where the model needs to identify if the table contains the experimental results of target models.}
    \label{tab:prompt_example}  
\end{table*}

\begin{table*}[ht]  
    \centering  
    \small
    \renewcommand{\arraystretch}{1.3} 
    \begin{tabular}{p{0.15\textwidth}|p{0.8\textwidth}}
    \toprule
    \textbf{Component} & \textbf{Prompt} \\ \midrule  

Context Augmentation & \texttt{Augment the extracted records from the table's LaTeX source by incorporating additional context from the text source to enrich and complete the records.}  

\texttt{Extracted Record Template: \{"value": "xx", "dataset": "xx", "dataset\_citation\_tag": "xx", "subset": "xx", "model\_name": "xx", "metric": "xx", "prompting\_method": "xx", "number\_of\_shots": "xx"\}}  \newline

\texttt{To accurately augment and enrich the extracted records, follow these steps systematically:}  

\texttt{1. value: Referencing the table source, if a numeric value is only partially extracted from a table cell, ensure that the entire content of the cell is used to update the value.}  

\texttt{2. dataset: Referencing the table source, table caption, and source text, locate the full name of the dataset and update the name accordingly.}  

\texttt{3. dataset\_citation\_tag: Referencing the table source, table caption, and source text, identify the dataset citation tag to the extracted record. Avoid using LaTeX syntax (e.g., cite and curly brackets); return only the tag name contained within.}  

\texttt{4. subset: Referencing the table source, table caption, and source text, identify specific subsets of the dataset, such as subtasks, domains, splits, or language pairs, and provide detailed descriptions of each subset. Prioritize the use of column information from the table source to identify the subset. If the subset is not explicitly mentioned in the table source, refer to the table caption or source text to identify the subset.}  

\texttt{5. model\_name: Referencing the table source, if a model name is partially extracted, revisit the corresponding table cell to ensure the entire content is included.}  

\texttt{6. metric: Referencing the table source, table caption, and source text, extract the metrics used in the experiment along with detailed descriptions and any additional information about the evaluation protocol.}  

\texttt{7. prompting\_method: Referencing the table source, table caption, and source text, search for and identify the prompting technique (e.g., direct, CoT, etc.) applied in the experiment and provide a detailed explanation of it. Do not include any information related to the number of shots (e.g., few-shot, zero-shot, three-shot) in this field.}  

\texttt{8. number\_of\_shots: Referencing the table source, table caption, and source text, specify the number of shots used in the experiment. This must be an integer value.}  \newline

\texttt{During output, output only the template following extracted results. Do not output any explanations or use LaTeX grammar. For any unanswerable attributes in the templates, leave their value as "xx".} 

\textbf{Input}

\texttt{Extracted Records: }  

\texttt{Table LaTeX Source: }  

\texttt{Text Source: }  

\textbf{Output}

\texttt{Augmented Extracted Records: }  \\
    \bottomrule
    \end{tabular}
    \caption{Prompt for context augmentation.}  
    \label{tab:prompt_example_context_augmentation}  
\end{table*}








\end{document}